\documentclass[10pt,twocolumn,letterpaper]{article}

\usepackage{iccv}              % To produce the CAMERA-READY version
\usepackage{amsmath}
\usepackage{amssymb}
\usepackage{mathtools}
\usepackage{pifont}
\usepackage{graphicx}
\usepackage{booktabs} % for professional tables
\usepackage{graphicx}
\usepackage{amsmath,bm}

\usepackage{xspace}
\usepackage{booktabs}
\usepackage{wrapfig}
\usepackage{multirow, multicol}
\usepackage{caption}
\usepackage{xcolor}
\usepackage{adjustbox}
\usepackage{cuted}
\usepackage{colortbl}

\usepackage{float}
\definecolor{iccvblue}{rgb}{0.21,0.49,0.74}
\usepackage[pagebackref,breaklinks,colorlinks,allcolors=iccvblue]{hyperref}
\title{TransAnimate: Taming Layer Diffusion to Generate RGBA Video}

\author{
Xuewei Chen\textsuperscript{$1$} \thanks{The first two authors contributed equally and are listed in alphabetical order.} \quad
Zhimin Chen\textsuperscript{$1$} \footnotemark[1] \quad
Yiren Song \textsuperscript{$2$} 
\\
$^1$ Clemson University.\quad 
$^2$ National University of Singapore.\quad 
}

\begin{document}
\maketitle
\vspace{-30pt} 
\begin{strip}
    \centering
        \vspace{-5em}

    \centering
    \includegraphics[width=0.93\textwidth]{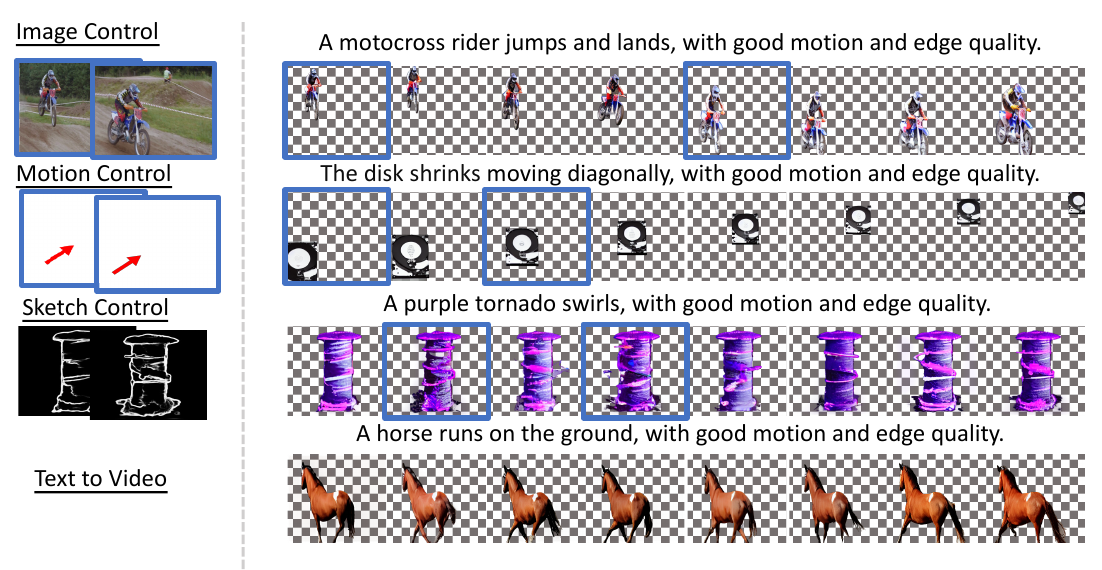}
    \captionof{figure}{\textbf{RGBA Video Generation with TransAnimate.} By utilizing pre-trained text-to-transparent image models, the motion-guided control mechanism, and the proposed dataset, TransAnimate enables high quality generation and effective control of video content.}
    \vspace{-10pt} 

    \label{fig:teaser}
\end{strip}
\begin{abstract}
Text-to-video generative models have made remarkable advancements in recent years. However, generating RGBA videos with alpha channels for transparency and visual effects remains a significant challenge due to the scarcity of suitable datasets and the complexity of adapting existing models for this purpose. To address these limitations, we present TransAnimate, an innovative framework that integrates RGBA image generation techniques with video generation modules, enabling the creation of dynamic and transparent videos. TransAnimate efficiently leverages pre-trained text-to-transparent image model weights and combines them with temporal models and controllability plugins trained on RGB videos, adapting them for controllable RGBA video generation tasks. Additionally, we introduce an interactive motion-guided control mechanism, where directional arrows define movement and colors adjust scaling, offering precise and intuitive control for designing game effects. To further alleviate data scarcity, we have developed a pipeline for creating an RGBA video dataset, incorporating high-quality game effect videos, extracted foreground objects, and synthetic transparent videos. Comprehensive experiments demonstrate that TransAnimate generates high-quality RGBA videos, establishing it as a practical and effective tool for applications in gaming and visual effects.
\end{abstract}    
\section{Introduction}
\label{sec:intro}

Text-to-video generation models have achieved remarkable progress in recent years, enabling the creation of dynamic and visually engaging content widely applied in video editing, image animation, and motion customization. Furthermore, methods incorporating control signals like optical flow, pose skeletons, and multimodal inputs have been proposed to achieve more precise video generation guidance. Alpha channels play a vital role in producing high-quality visual effects, as they allow transparent elements such as smoke, fire, and light to seamlessly integrate into complex scenes. This capability is particularly critical in game development, where transparency effects are central to creating immersive and realistic experiences. However, generating controllable RGBA videos with special effects remains highly desirable yet challenging.

Despite breakthroughs in video generation models, RGBA video synthesis remains underdeveloped. This dilemma stems from dual constraints: (1) the lack of large-scale RGBA video datasets severely limits algorithmic exploration, and (2) existing solutions like LayerDiffuse, while capable of generating static transparent images, cannot be seamlessly integrated with video generation models. The core challenge lies in leveraging massive RGB video pretraining resources and RGBA image generation models to build a data-efficient transparent video framework—an urgent problem we address.

This paper presents TransAnimate, a framework that unifies transparency modeling and video generation through three synergistic innovations. First, we establish an RGBA video synthesis pipeline and construct a foundational RGBA video dataset. Second, we combine AnimateDiff's motion modeling capabilities, LayerDiffuse's transparency generation expertise, and sparse-control video methods by fine-tuning adaptation layers on limited RGBA video data. Third, we design motion control mechanisms tailored for game visual effect artists, enabling pixel-precise control through directional arrows for trajectory specification and hue parameters for effect scaling.

While large-scale RGB video datasets contain millions of samples, the scarcity of RGBA video data remains a significant bottleneck. To address this limitation, we employ three complementary data collection strategies: (a) curating high-quality videos from game designers to capture authentic transparency properties, (b) extracting foreground object videos from instance segmentation data to enhance motion diversity, and (c) synthesizing controllable transparent videos using parametric transformations such as translation and scaling. Each strategy has distinct advantages and limitations: (a) provides superior visual quality but lacks category diversity, (b) introduces diverse motion patterns but suffers from incomplete foregrounds and imperfect edges, and (c) ensures precise edges and broad category coverage but is limited in motion complexity. Additionally, we manually curate high-quality results from models trained on these datasets and incorporate them as supplementary training data in an iterative refinement process. To balance these trade-offs, we propose a positive trigger strategy, which assigns distinct learnable tokens to each data source. This approach helps mitigate the impact of imperfect data distributions by preventing the model from absorbing unwanted features, thereby improving overall learning efficiency.

Architecturally, we first augment LayerDiffuse with temporal modules initialized using AnimateDiff's RGB-pretrained weights, eliminating costly training from scratch. Subsequent fine-tuning on RGBA video data enables effective adaptation of motion priors from AnimateDiff. For controllable generation, we repurpose RGB-pretrained SparseCtrl~\cite{guo2024sparsectrl} with input-layer fine-tuning—experiments show minimal parameter adjustments suffice for high-consistency control.

Addressing game developers' needs, we implement a vector-chroma joint control system: Motion vectorization enables 8-directional trajectory control with velocity parameterization, while chromatic scaling regulates effect magnitudes. This integrated control mechanism significantly enhances flexibility for game visual effects creation, empowering designers to produce high-quality customizable effects with unprecedented efficiency and precision. By unifying transparency modeling and motion dynamics, our method achieves diverse RGBA video generation while maintaining controllability comparable to RGB video approaches.

Our contributions are summarized as follows:
\begin{itemize}

\item We propose TransAnimate, a framework for generating temporally coherent layered transparent videos and effects.

\item We develop a dataset pipeline integrating game effects, segmented foregrounds, and synthetic videos, refined via iterative enhancement and a positive trigger strategy to mitigate distribution mismatches.

\item We introduce novel directional-chromatic controls enabling pixel-accurate motion specification for game effects.
\end{itemize}

\section{Related Work}
\label{sec:re}

\begin{figure*}[t!]
\centering
\includegraphics[width=0.9\linewidth]{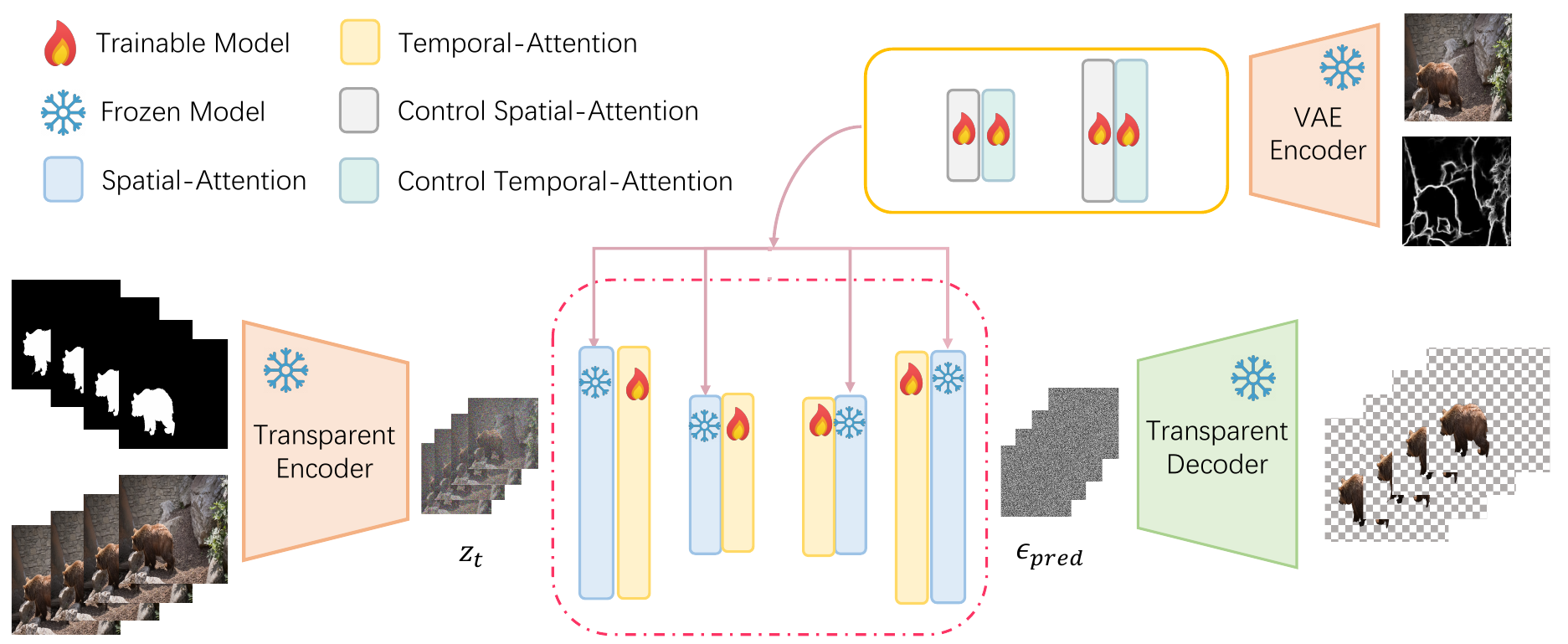}
\caption{\textbf{Framework Overview.} TransAnimate generates transparent videos by learning motion patterns from videos. A frozen Transparent Encoder extracts features, refined by Temporal Attention and Linear Layers. Pre-trained SparseCtrl weights enable control via motion, sketches, and RGB images. A frozen Transparent Decoder reconstructs transparent frames, enhancing generation with limited RGBA data.}
\vspace{-7pt}
\label{fig:framework}
\end{figure*}

\noindent \textbf{Text-to-video generation.}
Recent advancements in text-to-video (T2V) generation~\cite{karras2023dreampose, ruan2023mm, zhang2023i2vgen, he2022latent, chen2023videocrafter1, hong2022cogvideo} have predominantly utilized diffusion models~\cite{ho2020denoising, song2020denoising, peebles2023scalable, ren2024mushroom}, celebrated for their training stability and robust open-source ecosystems. A foundational milestone in this domain is the Video Diffusion Model~\cite{ho2022video}, which extends a 2D image diffusion framework to handle video data by jointly training on image and video datasets from scratch. Building upon this, subsequent approaches leverage pre-trained image generators, such as Stable Diffusion~\cite{rombach2022high}, by integrating temporal layers into the existing 2D architectures and fine-tuning on expansive video datasets~\cite{bain2021frozen}. Among innovative methods, Align-Your-Latents~\cite{blattmann2023align} achieves efficient T2V conversion by aligning noise maps sampled independently for each frame, while AnimateDiff~\cite{animatediff} incorporates a modular motion layer, enabling the generation of high-quality animations on customized image generation backbones~\cite{ruiz2023dreambooth}. To address temporal coherence challenges, Lumiere~\cite{bar2024lumiere} eliminates the need for a temporal super-resolution module by directly generating videos with consistent frame rates. Further notable advancements include adopting scalable transformer architectures~\cite{ma2024latte}, leveraging spatiotemporal compressed latent spaces, as demonstrated by W.A.L.T.\cite{gupta2023photorealistic} and Sora\cite{videoworldsimulators2024}, and utilizing discrete tokens alongside language models for video synthesis, exemplified by VideoPoet~\cite{kondratyuk2023videopoet}. Building on previous RGB video generation approaches, our method introduces text-to-RGBA video generation tailored for game effects design, addressing a significant gap in this domain.

\noindent \textbf{Controllable video generation.}
Existing text-to-video models often suffer from limited control, as relying solely on text descriptions introduces ambiguity and reduces precision. To address this, various methods incorporate explicit guidance signals. For instance, some approaches use depth maps or skeleton sequences to dictate scene layout or human motion with greater accuracy~\cite{sparsectrl, chen2023control, zhang2023controlvideo, hu2023animate, turkulainen2024dn, chen2023bridging}. Others leverage image-based control signals, which enhance video quality and improve temporal consistency~\cite{sparsectrl, conditionalimage}. Despite these advances, controlling camera motion during video generation remains underexplored. AnimateDiff~\cite{animatediff} applies LoRA~\cite{hu2021lora} fine-tuning to adapt model weights for specific camera angles. Direct-a-Video~\cite{directavideo} incorporates a camera embedder for pose control but supports only basic parameters, limiting its capacity to simple motions like panning. MotionCtrl~\cite{motionctrl} extends this idea with additional input parameters to enable complex camera trajectories. However, its reliance on fine-tuning components of the diffusion model compromises generalization. In this work, we introduce a novel motion-guided control framework specifically designed for generating transparent videos with predefined motion directions and scales. This method fills a critical gap by offering game effect designers a powerful and flexible tool for creating dynamic, high-quality content that meets the unique demands of their workflows.

\begin{figure*}[t!]
\includegraphics[width=\linewidth]{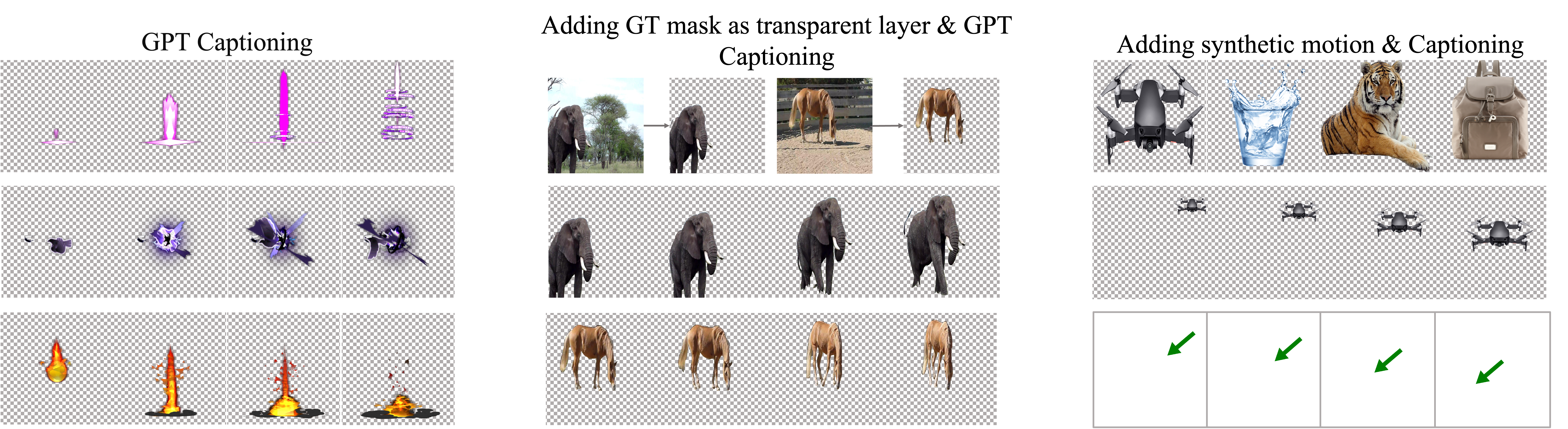}
\begin{tabular}{p{4.7cm} p{5.5cm} p{6cm}}
        \centering (a). Game effect dataset & \centering (b). Foreground dataset & \centering (c). Synthesized motion dataset
\end{tabular}
\vspace{-15pt}
\caption{ \textbf{ Illustration of TransAnimate.} Our dataset consists of (a) Animate Dataset with 3,000 high-quality game effect videos, (b) Foreground Object Videos Dataset with 7,000 segmented videos capturing diverse motion patterns, and (c) Synthesized Transparent Motion Videos with 20,000 generated samples featuring controlled motion transformations. For synthesized dataset, from top to bottom, it represents Motion Caption, Raw Image, Synthetic, and Motion Control.}
\label{fig:framework}
\vspace{-15pt} 
\end{figure*}
\textbf{Transparent Layer Processing}
Most existing approaches for transparent layer generation focus primarily on image generation, closely linked to image matting techniques. For instance, PPMatting ~\cite{chen2022pp} is a neural network model for image matting, trained from scratch using standard matting datasets. Building on advancements in foundational models, Matting Anything ~\cite{li2024matting} leverages the Segment Anything Model (SAM)~\cite{kirillov2023segment} as its backbone for matting tasks, while VitMatte ~\cite{yao2024vitmatte} utilizes a Vision Transformer (ViT) in a tri-map-based matting framework. Expanding beyond traditional matting, recent innovations explore the integration of layered effects in diffusion models. LayerDiffuse~\cite{zhang2024transparent} introduces the concept of "latent transparency." This technique encodes alpha channel transparency directly into the latent manifold of a pre-trained latent diffusion model by modifying the Variational Autoencoder (VAE) to decode alpha channels, enabling richer transparency effects. Despite these advancements, transparent video generation remains significantly underexplored. SAM-2~\cite{ravi2024sam2} integrates a Transformer architecture with streaming storage and memory mechanisms to provide coherent segmentation predictions across video sequences. However, SAM-2 is not explicitly designed for video generation and struggles to create diverse, high-quality transparent layers essential for effects like fire, light, and smoke, which are crucial in game effect design. In this work, we introduce a novel framework tailored for game effect generation, enabling text-to-RGBA video generation with support for various control modalities. Our approach addresses the limitations of existing methods, providing an innovative solution for creating visually compelling and dynamic transparent layers in video content, specifically for gaming applications.

\section{Method}

Our approach enables a latent diffusion model to learn transferable motion priors from video data while leveraging existing text-to-transparent-image generation methods to produce transparent videos. In Section~\ref{sec:preliminary}, we introduce the foundational principles of our method. Section~\ref{sec:Dataset} details the dataset preparation process. Section~\ref{sec:positive_prompts} demonstrates how the proposed positive prompts enhance motion and edge quality. Section~\ref{sec:TransAnimate} elaborates on the TransAnimate framework for transparent video generation. Finally, Section~\ref{sec:Control} explores how pretrained SparseCtrl weights are adapted to achieve controllable transparent video generation.

% 我们的方法使潜在扩散模型（LDM）能够从视频数据中学习可迁移的运动先验，这一方法基于现有的文本到透明图像生成方法，用于产生透明视频。在第\ref{sec:preliminary}节中，我们介绍了我们方法的基本原理。第\ref{sec:TransAnimate}节详细介绍了我们提出的透明视频生成框架。在第\ref{sec:Control}节中，我们扩展了预训练的SparseCtrl权重的功能，使其能够实现可控的透明视频生成。最后，在第\ref{sec:Dataset}节中，我们概述了数据集准备过程。

\subsection{Preliminary}\label{sec:preliminary}

\begin{figure*}[t!]
\centering
\includegraphics[width=0.9 \linewidth]{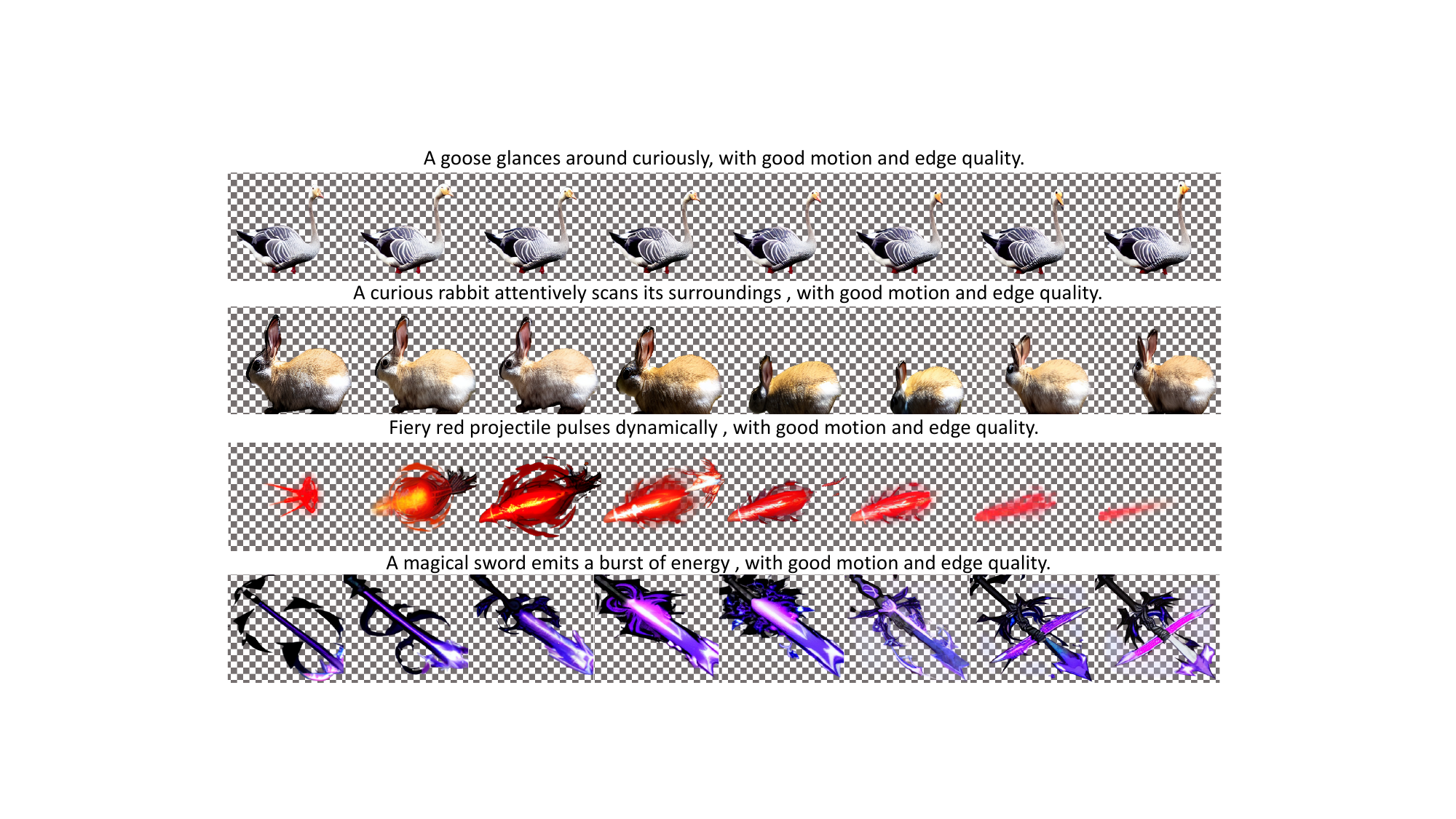}
\vspace{-8pt}
\caption{ Text-to-RGBA video generation results of TransAnimate. }
\label{fig:video}
\vspace{-13pt}
\end{figure*}
Stable Diffusion (SD), a widely-used open-source text-to-image (T2I) model, performs the diffusion process in the latent space of a pre-trained autoencoder. This process perturbs the encoded image \( z_0 = \mathcal{E}(x_0) \) into \( z_t \) at step \( t \) by adding noise:  
\begin{equation}
z_t = \sqrt{\bar{\alpha_t}}z_0 + \sqrt{1-\bar{\alpha_t}}\epsilon,~\epsilon \sim \mathcal{N}(0, \mathit{I}),
\end{equation}  
where \( \bar{\alpha_t} \) governs the noise strength. The denoising UNet \( \epsilon_\theta \), utilizing an MSE loss, predicts noise while integrating ResNet, self-attention, and cross-attention mechanisms to incorporate text conditions effectively.

AnimateDiff~\cite{animatediff} is an extension of T2I models, leverages video data to learn transferable motion priors, enabling animation generation through iterative denoising. The method consists of three core components: a domain adapter, a motion module, and a LoRA-based extension, all seamlessly integrated into the T2I framework during inference.

To model temporal dynamics, AnimateDiff inflates the 2D T2I model to handle 3D video data. Video tensors \( x \in \mathbb{R}^{b \times c \times f \times h \times w} \) are reshaped to allow the image layers to process frames independently, while a newly introduced motion module captures temporal dependencies across frames.

The motion module in AnimateDiff employs a temporal Transformer enhanced with self-attention blocks and sinusoidal position encodings. Input features are reshaped into sequences along the temporal axis to facilitate inter-frame information exchange:  
\begin{equation}
z_{out} = \text{Softmax}(QK^T / \sqrt{c}) \cdot V,
\end{equation}  
where \( Q, K, V \) are linear projections of the input features. This design effectively encodes motion priors, enabling the generation of coherent and realistic animations.

% Stable Diffusion（SD）是一个广泛使用的开源文本到图像（T2I）模型，它在预训练的自动编码器的潜在空间中执行扩散过程。该过程通过添加噪声将编码后的图像 \( z_0 = \mathcal{E}(x_0) \) 扰动为第 \( t \) 步的 \( z_t \)：
% \begin{equation}
% z_t = \sqrt{\bar{\alpha_t}}z_0 + \sqrt{1-\bar{\alpha_t}}\epsilon,~\epsilon \sim \mathcal{N}(0, \mathit{I}),
% \end{equation}
% 其中 \( \bar{\alpha_t} \) 控制噪声强度。去噪UNet \( \epsilon_\theta \)，使用MSE损失，预测噪声的同时，整合了ResNet、自注意力和交叉注意力机制，以有效地整合文本条件。

% AnimateDiff~\cite{animatediff} 是T2I模型的扩展，利用视频数据学习可迁移的运动先验，通过迭代去噪生成动画。该方法由三个核心组件组成：域适配器、运动模块和基于LoRA的扩展，所有这些在推理期间无缝集成到T2I框架中。

% 为了建模时间动态，AnimateDiff将2D T2I模型扩展到处理3D视频数据。视频张量 \( x \in \mathbb{R}^{b \times c \times f \times h \times w} \) 被重塑以允许图像层独立处理帧，同时新引入的运动模块捕获帧间的时间依赖性。

% AnimateDiff中的运动模块使用增强的自注意力块和正弦位置编码的时间Transformer。输入特征沿时间轴重塑成序列，以促进帧间信息交换：
% \begin{equation}
% z_{out} = \text{Softmax}(QK^T / \sqrt{c}) \cdot V,
% \end{equation}
% 其中 \( Q, K, V \) 是输入特征的线性投影。这种设计有效地编码了运动先验，使得生成的动画连贯且真实。

\subsection{New Dataset for Transparent Video Generation} \label{sec:Dataset}

Achieving high-quality transparent video generation presents three major challenges: (1) ensuring high-fidelity transparent layers, (2) capturing realistic and dynamic motion, and (3) maintaining sufficient diversity in video content. To address these challenges, we construct a comprehensive dataset by collecting and synthesizing three distinct types of video data, ensuring that our network learns from diverse samples with high-quality motion and transparency.

\textbf{Animate Dataset.} The animate dataset consists of 3,000 professionally curated game effect videos created by expert designers. These videos feature complex and visually rich animations, such as explosions, energy transformations, and magical effects. Each video is carefully selected to ensure high transparency quality and realistic motion.

\begin{table}[t!]
    \centering
    \small
    \renewcommand{\arraystretch}{1.2} % Increase row spacing
    \setlength{\tabcolsep}{4pt} % Adjust column spacing
    \begin{tabular}{lccc}
        \toprule
        \multirow{2}{*}{Dataset} & RGBA Edge & Motion & Category \\
        & Quality & Quality & Diversity \\
        \midrule
        Game Effects Dataset & \checkmark & \checkmark & \ding{55} \\
        Foreground Dataset & \ding{55} & \checkmark & \checkmark \\
        Synthetic Dataset & \checkmark & \ding{55} & \checkmark \\
        Iterated Dataset & \checkmark & \checkmark & \checkmark \\
        \bottomrule
    \end{tabular}
    \caption{Comparison of datasets based on three key quality aspects.}
    \vspace{-15pt}
    \label{table:dataset}
\end{table}

\textbf{Foreground Object Videos Dataset.} To enhance diversity and capture a wide range of motion patterns, we construct a foreground object dataset by extracting 7,000 videos from large-scale video instance segmentation datasets, including VideoMatte240K~\cite{lin2021real}, MeViS~\cite{ding2023mevis}, YouTube-VOS~\cite{xu2018youtube}, DAVIS~\cite{pont20172017}, and MOSE~\cite{ding2023mose}. These datasets cover a broad range of object categories, ensuring the dataset includes diverse appearances, textures, and motion trajectories. Using video instance segmentation techniques, we extract foreground objects and retain only those exhibiting significant motion dynamics.

\begin{figure*}[t!]
\centering
\includegraphics[width=0.9 \linewidth]{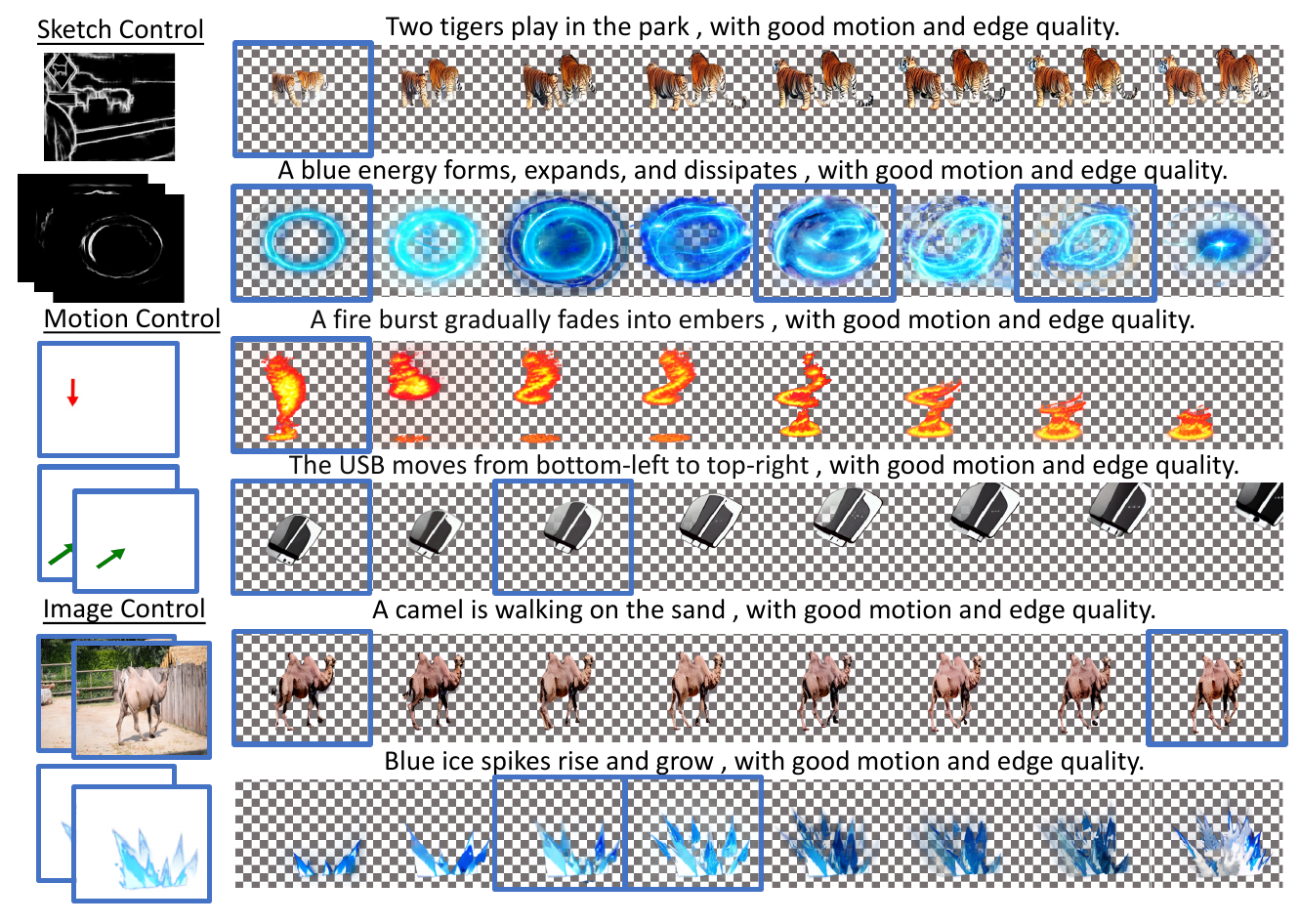}
\vspace{-10pt}
\caption{  Conditional generation results from TransAnimate. The qualitative results are results with sketch, depth, and RGB image conditions. The input conditions are displayed on the left, while the keyframes guided by these conditions are highlighted with blue borders.}
\vspace{-12pt}
\label{fig:control}
\end{figure*}

\textbf{Synthesized Transparent Motion Videos.} To further improve transparency quality and motion diversity, we generate 20,000 synthetic transparent videos by applying controlled motion transformations to foreground images from predefined transparent classes. These transformations include translation, scaling, and rotation, ensuring that the dataset captures a wide variety of movement behaviors. The motion characteristics are visualized with directional arrows (for translation) and color-coded indicators (green for scaling up, blue for stability, and red for shrinking). Captions are generated based on the transparent class and applied motion transformation, ensuring a strong alignment between textual descriptions and visual content.

\textbf{Data Iteration} We jointly train the TransAnimate model on the aforementioned three datasets, employing a data iteration strategy to manually curate high-quality generated results as supplementary training data. This effectively mitigates the scarcity of RGBA data. The animate dataset ensures high-quality transparent layers and realistic motion, the foreground object dataset provides diversity in motion and appearance, and the synthesized transparent motion dataset further refines transparency and motion control. To address the limitations of different datasets, we adopt a negative trigger strategy to filter out undesirable features, significantly enhancing the performance of transparent video generation.

% \textbf{Final Dataset Compilation.} By combining these three datasets, we construct a comprehensive dataset of 30,000 videos that effectively addresses the core challenges of transparent video generation. The animate dataset ensures high-quality transparent layers and realistic motion, the foreground object dataset provides motion and appearance diversity, and the synthesized transparent motion dataset further refines transparency and motion control. With this dataset, our framework is capable of learning a diverse, motion-rich, and transparency-aware representation, significantly improving transparent video generation performance.
% % 

% \subsection{透明视频生成的新数据集} \label{sec:Dataset}

% 实现高质量的透明视频生成面临三大挑战：(1) 确保高保真透明层，(2) 捕捉逼真且动态的运动，(3) 保持视频内容的足够多样性。为了解决这些挑战，我们通过收集和合成三种不同类型的视频数据构建了一个全面的数据集，确保我们的网络能够从具有高质量运动和透明度的多样化样本中学习。

% \textbf{动画数据集。} 动画数据集由 3,000 个专业策划的游戏特效视频组成，这些视频来自专业设计师。这些视频包含复杂且视觉效果丰富的动画，例如爆炸、能量转换和魔法效果。每个视频都经过精心挑选，以确保高透明质量和合理的运动。

% \textbf{前景对象视频数据集。} 为了增强多样性并捕捉广泛的运动模式，我们通过从大规模视频实例分割数据集中提取 7,000 个视频构建了一个前景对象数据集，这些数据集包括 VideoMatte240K、MeViS、YouTube-VOS、DAVIS 和 MOSE。这些数据集涵盖了广泛的物体类别，确保数据集包含多样化的外观、纹理和运动轨迹。利用视频实例分割技术，我们提取前景对象并仅保留那些表现出显著运动动态的视频。

% \textbf{合成的透明运动视频。} 为了进一步提高透明质量和运动多样性，我们通过对预定义透明类别的前景图像应用受控运动变换，生成了 20,000 个合成透明视频。这些变换包括平移、缩放和旋转，确保数据集捕捉到各种运动行为。运动特性通过方向箭头（用于平移）和颜色编码指示器（绿色表示放大，蓝色表示稳定，红色表示缩小）进行可视化。标题根据透明类别和应用的运动变换生成，确保文本描述与视觉内容之间的强一致性。

% \textbf{数据迭代。} 我们在上述三个数据集上联合训练Text2RGBA模型， 采用数据迭代策略人工筛选高质量生成结果作为训练集的补充，有效缓解了RGBA数据匮乏的问题。动画数据集确保了高质量的透明层和逼真的运动，前景对象数据集提供了运动和外观的多样性，而合成的透明运动数据集进一步优化了透明度和运动控制， 对于不同数据集的劣势， 我们用negative trigger策略吸收不想要的特征，显著提升了透明视频生成的性能。

\subsection{Positive  Triggers} 
\label{sec:positive_prompts}
As shown in Table ~\ref{table:dataset}, the datasets we collect exhibit certain limitations in terms of quality attributes. Specifically, the Foreground Dataset demonstrates strong motion quality but suffers from poor edge quality, while the Synthetic Dataset excels in edge quality but lacks sufficient motion fidelity. To mitigate these issues, we leverage positive triggers during training, explicitly guiding the model to learn high-quality attributes from different datasets.

For instances originating from the Synthetic Dataset, we include positive triggers emphasizing good edge quality, helping the model recognize and reinforce well-defined edges. Conversely, for samples from the Foreground Dataset, we incorporate positive triggers highlighting strong motion attributes, enabling the model to better capture dynamic motion patterns. This targeted approach allows the network to develop a more comprehensive understanding of both motion and edge quality during training.

During inference, we combine positive triggers for both good edge and motion quality, ensuring that the generated results exhibit improved sharpness and dynamic consistency. This strategy enhances the overall quality of RGBA video generation by bridging the strengths of multiple datasets.

\subsection{Text2RGBA Video Generation} \label{sec:TransAnimate}

Our method introduces a novel approach to RGBA video generation by extending text-to-transparent-image frameworks, particularly by using LayerDiffuse~\cite{zhang2024transparent}. Unlike prior methods focused solely on RGB, we address the unique requirements of generating transparent animations, crucial for game effects design.

The framework integrates a motion module inspired by AnimateDiff~\cite{animatediff}, enabling animation through iterative denoising. This module is trained independently while keeping the text-to-transparent-image generation network frozen. By incorporating this learnable module into LayerDiffuse, we achieve temporal animation while preserving the high-quality transparency features of the original model.

To adapt LayerDiffuse for video generation, we extend its architecture to process 5D video tensors $\bm{x} \in \mathbb{R}^{b \times c \times f \times h \times w}$, where $b$, $c$, $f$, $h$, and $w$ denote the batch size, channels, frames, height, and width, respectively. Following a network inflation strategy similar to AnimateDiff~\cite{blattmann2023align}, we allow the image layers to process frames independently by reshaping the temporal axis $f$ into the batch axis during feature extraction and restoring it afterward.

Conversely, the motion module reshapes spatial dimensions ($h, w$) into the batch axis during temporal processing, which is restored post-processing. This design ensures efficient and independent handling of spatial and temporal dimensions.

% \begin{figure*}[t!]
% \centering
% \includegraphics[width=0.95 \linewidth]{Figures/abl_control.png}
% \caption{ \textbf{ Illustration of TransAnimate.} }
% \label{fig:ablation}
% \end{figure*}

% \begin{figure*}[t!]
% \centering
% \includegraphics[width=0.95 \linewidth]{Figures/abl_dataset.png}
% \caption{ \textbf{ Illustration of TransAnimate.} }
% \label{fig:ablation_dataset}
% \end{figure*}
% \begin{figure*}[t!]
%     \centering
%     % Figure 6a
%     \subfigure[Effect of the trained control network.]{
%         \includegraphics[width=0.48\linewidth]{Figures/Image6a.png}
%         \label{fig:transanimate1}
%     }
%     \hfill
%     % Figure 6b
%     \subfigure[One-stage and two-stage datasets.]{
%         \includegraphics[width=0.48\linewidth]{Figures/Image6b.png}
%         \label{fig:transanimate2}
%     }
%     \caption{\textbf{Illustration of TransAnimate.} (a) Trained control network. (b) One-stage vs. two-stage dataset.}
%     \label{fig:transanimate}
% \end{figure*}
\begin{figure*}[t!]
    \centering
    \begin{minipage}{0.48\linewidth}
        \centering
        \includegraphics[width=\linewidth]{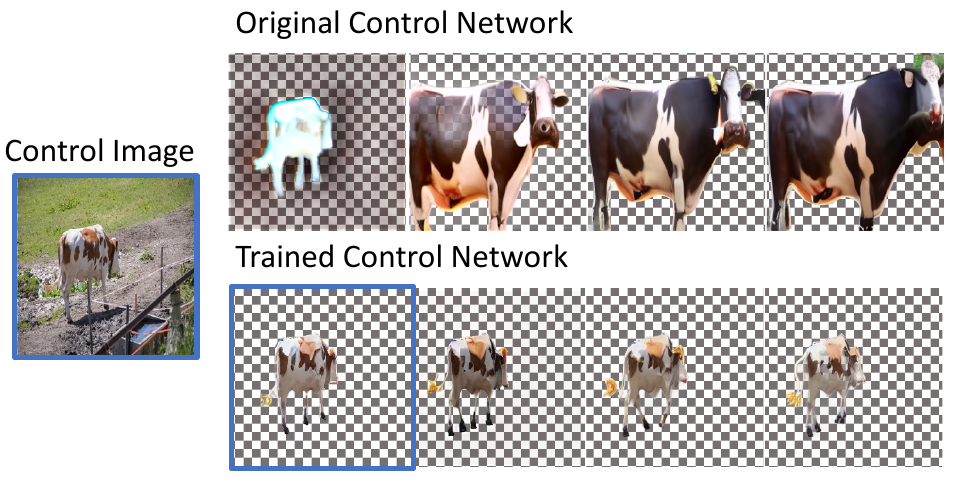}
        \caption{Qualitative comparison of SparseControl's original weights with our trained weights.}
        \vspace{-7pt}
        \label{fig:qual_control}
    \end{minipage}%
    \hfill
    \begin{minipage}{0.48\linewidth}
        \centering
        \includegraphics[width=\linewidth]{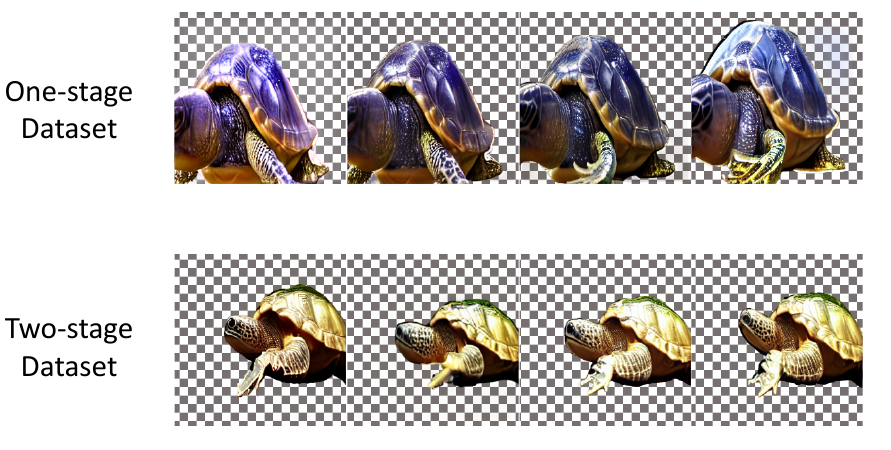}
        \caption{Qualitative comparison of results obtained using one-stage and two-stage training datasets.}
        \vspace{-7pt}
        \label{fig:qual_data}
    \end{minipage}
    \label{fig:example}
\end{figure*}

\begin{table*}[t!]
    \centering
    \begin{minipage}{0.48\linewidth}
        \centering
        \begin{tabular}{lcccc}
            \toprule
            & \multirow{2}{*}{FVD} & \multirow{2}{*}{CLIP} & RGBA & Motion \\
            &  &  & Quality & Quality \\
            \midrule
            Original control & 1400.86 & 19.22 & 16.32 & 18.17 \\
            \rowcolor{gray!8}Trained control  & \textbf{$\mathbf{407.41}$} & \textbf{$\mathbf{30.57}$} & \textbf{$\mathbf{83.62}$} & \textbf{$\mathbf{82.28}$} \\
            \bottomrule
        \end{tabular}
    \caption{Quantitative comparison of SparseControl's original weights with our trained weights.}
        \vspace{-10pt}
        \label{table:quant_control}
    \end{minipage}%
    \hfill
    \begin{minipage}{0.48\linewidth}
        \centering
        \begin{tabular}{lcccc}
            \toprule
            & \multirow{2}{*}{FVD} & \multirow{2}{*}{CLIP} & RGBA & Motion \\
            &  &  & Quality & Quality \\
            \midrule
            One-stage  & 485.63 & 28.55 & 67.05 & 69.17 \\
            \rowcolor{gray!8}Two-stage  & \textbf{$\mathbf{429.82}$} & \textbf{$\mathbf{29.32}$} & \textbf{$\mathbf{76.32}$} & \textbf{$\mathbf{78.79}$} \\
            \bottomrule
        \end{tabular}
        \caption{Quantitative comparison of results obtained using one-stage and two-stage training datasets.}
        \vspace{-10pt}
        \label{table:quant_dataset}
    \end{minipage}

\end{table*}

The motion module captures temporal dependencies by dividing the reshaped feature map along the temporal axis into a sequence of vectors $\{z_1, z_2, \dots, z_f\}$, where each vector corresponds to a frame. These vectors are processed through self-attention layers, enabling temporal interactions:
\vspace{-10pt}
\begin{equation}
    z_{out} = \text{Softmax}\left(\frac{QK^\top}{\sqrt{c}}\right)V,
\end{equation}
where $Q = W^Qz$, $K = W^Kz$, and $V = W^Vz$ are the query, key, and value projections of the input features. This mechanism ensures temporal coherence by enabling information flow across frames.

During training, the encoder, decoder, and 3D U-Net components of LayerDiffuse are frozen, and only the motion module is trained. The diffusion process progressively adds noise to the latent representation $\bm{x}_t$ over time steps $t$. The network predicts the added noise using a denoising model $\epsilon_\theta$, minimizing the following objective:
\begin{equation}
    \mathcal{L} = \mathbb{E}_{\mathcal{E}(x_0^{1:f}), y, \epsilon^{1:f} \sim \mathcal{N}(0, I), t} \left[\lVert \epsilon - \epsilon_\theta(z_t^{1:f}, t, \tau_\theta(y)) \rVert_2^2 \right].
\end{equation}

At inference, the pre-trained latent transparency decoder of LayerDiffsion, $\mathcal{D}(\cdot, \cdot)$, reconstructs the video’s RGB and alpha channels. Given the adjusted latent representation $\bm{x}_a$, the decoder outputs the transparent video:
\begin{equation}
    [\hat{\bm{I}_c} \,\, \hat{\bm{I}_\alpha}] = \mathcal{D}(\hat{\bm{I}}, \bm{x}_a),
\end{equation}
where $\hat{\bm{I}_c}$ and $\hat{\bm{I}_\alpha}$ represent the reconstructed color and alpha channels, respectively. This process ensures high-quality RGBA video outputs suitable for game effects.

\subsection{Controllable RGBA Video Generation} \label{sec:Control}

This section demonstrates how to apply SparseCtrl~\cite{sparsectrl}, which was trained on RGB videos, to RGBA controllable generation, incorporating three types of controls: motion control, RGB image control, and sketch control. Notably, we propose utilizing only the RGB channels of RGBA videos as control inputs, which is more practical for downstream tasks. We reuse the pre-trained weights of SparseCtrl and keep them frozen, while fine-tuning the Adapter layer on RGBA videos to align with TransAnimate Unet. This strategy effectively leverages the prior knowledge of the pre-trained RGB model, enabling robust RGBA video generation and control even with limited RGBA training data.

\textbf{Motion-Guided Video Generation.} Motion guidance is crucial for intuitive game effect design, enabling creators to precisely control the direction and scale of animations. Existing methods often lack support for such fine-grained functionality. Our motion control mechanism allows users to define the generation direction using arrows and indicate the effect scale using colors. For example, a transparent video effect can be guided to move from left to right while gradually shrinking, providing game designers with enhanced control and adaptability for their creative workflows.

\textbf{Sketch-to-Video Generation.} Sketches offer an intuitive and accessible way for non-professional users to guide T2V generation. SparseCtrl~\cite{sparsectrl} allows users to input sketches to shape video content. A single sketch can define the overall layout, while multiple sketches—such as for the first, last, and key intermediate frames—can guide coarse motion and transitions. This functionality is especially suited for tasks like storyboarding, enabling creators to visualize and iterate on video concepts effortlessly.

\textbf{Image Animation and Transition.} SparseCtrl ~\cite{sparsectrl} unifies various video generation tasks, including video prediction, animation, and interpolation, under a single framework leveraging RGB image conditions. Image animation generates videos based on the first frame, while transitions are guided by both the first and last frames. Video prediction uses initial frames to extrapolate motion, and interpolation creates smooth transitions between sparsely provided keyframes. By unifying these tasks, SparseCtrl broadens the applicability of video generation methods, making them versatile tools for diverse creative and practical scenarios.

\section{Experiments}
\subsection{Experiment Setting}
\noindent \textbf{Setup.} Our method leverages the pre-trained VAE encoder, decoder, and 3D-UNet architecture from LayerDiffuse, with the motion module adapted from AnimateDiff. The training resolution is set to \(256 \times 256\) for 16 frames. We train the model for a total of 3,000 iterations with a batch size of 16, utilizing two NVIDIA A100 80GB GPUs. The learning rate is set to \(1 \times 10^{-5}\), ensuring stable and effective optimization.

\noindent \textbf{Evaluation Metrics.} We evaluate RGBA and motion quality using CLIP Score ~\cite{radford2021learning}, FVD~\cite{unterthiner2019fvd}, and a user study. In the user study, 15 participants assess 50 generated videos based on two key aspects: (1) the quality of RGBA edges and (2) the realism and smoothness of motion.

\subsection{Qualitative Results}
Figure ~\ref{fig:video} presents qualitative results. These examples illustrate the model’s ability to produce high-quality transparency effects while preserving fine details and structural consistency. Furthermore, the results demonstrate the model’s strong generalization across various content types, reinforcing its robustness and adaptability.
% \begin{figure*}[t!]
% \centering
% \includegraphics[width=0.95 \linewidth]{Figures/TransmittAblation 2.pdf}
% \vspace{-10pt}
% \caption{ Qualitative comparison of results with and without the proposed positive triggers. }
% \vspace{-5pt}
% \label{fig:qual_prompt}
% \end{figure*}
\subsection{Qualitative Controllable Results}
Our approach seamlessly integrates with SparseControl to facilitate multi-modal video generation, enabling precise control over object structure, movement, and appearance. Users can provide sketches, motion trajectories, or reference images to guide the generation process. Figure ~\ref{fig:control} illustrates qualitative results, demonstrating the model’s capability to generate transparent images with high fidelity while maintaining user-defined control inputs. These results highlight the model’s ability to produce visually compelling outputs and generalize effectively across diverse content.

% \begin{table*}[t!]
%     \centering
%     \begin{tabular}{lccccc}
%         \toprule
%       & FID (↓)  & FVD (↓) & CLIP (↑) & RGBA Quality(↑) & Motion Quality (↑) \\
%         \midrule
%        Ours & 41.26 & 1400.86 & 19.22 & 16.32 & 18.17 \\
%        Ours & 15.83 & 307.41 & 30.57 & 83.62 & 82.28 \\
%         \bottomrule
%     \end{tabular}
%     \caption{Quantitative results}
%     \label{table:quant}

% \end{table*}

% \begin{table*}[t!]
%     \centering
%     \begin{tabular}{lcccc}
%         \toprule
%         & FVD  & CLIP& RGBA Quality & Motion Quality  \\
%         \midrule
%        Original control & 1400.86 & 19.22 & 16.32 & 18.17 \\
%        Trained control  & 307.41 & 30.57 & 83.62 & 82.28 \\
%         \bottomrule
%     \end{tabular}
%     \caption{Quantitative results}
%     \label{table:quant}
% \end{table*}

\begin{figure}[t!]
\centering
\includegraphics[width=0.95 \linewidth]{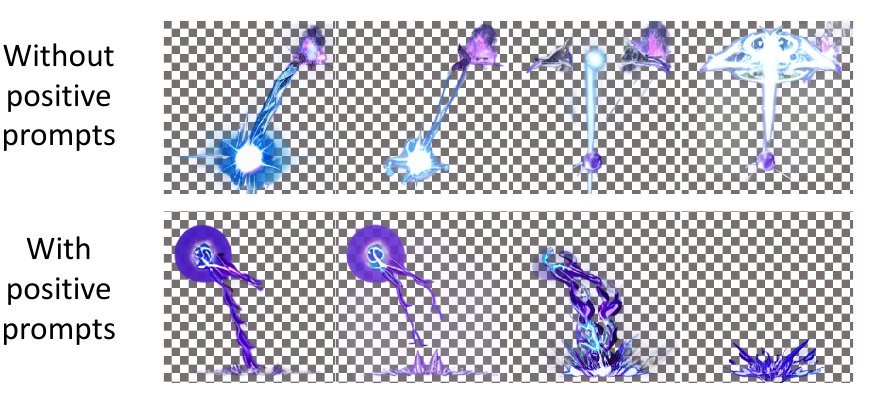}
\caption{ Qualitative comparison of results with and without the proposed positive triggers.}
\vspace{-10pt}
\label{fig:qual_prompt}
\end{figure}

\subsection{Ablative Study}

\textbf{Effectiveness of Control Training.} We evaluate the effectiveness of our trained control method by comparing it with the pre-trained weights provided by SparseControl. As shown in Figure ~\ref{fig:qual_control} and Table ~\ref{table:quant_control}, directly applying RGB pre-trained weights fails to preserve structural consistency in controllable RGBA video generation. In contrast, our approach successfully adapts RGB-based control to RGBA videos, achieving superior alignment with the control image while maintaining transparency effects. These results highlight the ability of pre-trained RGB control weights to generalize effectively to RGBA video generation, addressing the challenges posed by the scarcity of RGBA training data.

\textbf{Two-Stage Training.} We examine the impact of a two-stage training strategy on generation quality. As illustrated in Figure ~\ref{fig:qual_data}, this approach enhances dataset diversity and scalability without compromising visual fidelity, resulting in more stable and high-quality video generation. Additionally, the quantitative results in Table ~\ref{table:quant_dataset} confirm that the two-stage training strategy significantly improves the final performance of the generated RGBA videos, demonstrating its effectiveness in refining output quality.

\textbf{Positive Triggers.} We analyze the contribution of the proposed positive prompts strategy in enhancing generation quality, particularly in motion coherence and RGBA edge fidelity. As shown in Figure ~\ref{fig:qual_prompt}, this technique enriches dataset diversity while preserving structural integrity, leading to more consistent and visually refined outputs. Furthermore, Table ~\ref{table:quant_prompt} provides quantitative evidence that incorporating positive prompts significantly enhances the overall quality of generated RGBA videos, reinforcing its effectiveness as a control mechanism.

\begin{table}[t!]
    \centering
    \resizebox{0.46\textwidth}{!}{
        \begin{tabular}{lcccc}
            \toprule
            & \multirow{2}{*}{FVD} & \multirow{2}{*}{CLIP} & RGBA & Motion \\
            &  &  & Quality & Quality \\
            \midrule
            w/o Positive Prompts & 458.37 & 28.95 & 71.58 & 73.31 \\
            \rowcolor{gray!8}w Positive Prompts  & \textbf{$\mathbf{429.827}$} &\textbf{$\mathbf{29.32}$} & \textbf{$\mathbf{76.32}$} & \textbf{$\mathbf{78.79}$} \\        
            \bottomrule
        \end{tabular}
    }
    \caption{Quantitative comparison of results with and without the proposed positive triggers.}
    \vspace{-15pt}
    \label{table:quant_prompt}
\end{table}

\section{Conclusion}

In this paper, we address the underexplored challenge of RGBA video generation by introducing TransAnimate, a novel framework that combines transparency modeling with motion dynamics to generate high-quality, layered, transparent videos with temporal coherence. To tackle the scarcity of RGBA datasets, we propose a three-step data creation strategy, leveraging high-quality game effect videos, extracted foreground objects, and synthetic transparent videos with controlled motion dynamics. We also introduce a motion-guided control mechanism that enables precise adjustments to motion direction and scaling, tailored to game effect creation. By utilizing RGB channels as control inputs, our approach effectively leverages pre-trained RGB models, enabling robust RGBA video generation and control with limited RGBA data. Our contributions bridge critical gaps in RGBA video generation, providing a unified framework with controllability comparable to RGB methods while extending capabilities to layered transparency. This work opens new possibilities for immersive and customizable content creation in gaming and visual effects.

{
    \small
    \bibliographystyle{ieeenat_fullname}
    \bibliography{main}
}

\end{document}